\documentclass[runningheads]{llncs}
\usepackage{graphicx}

\usepackage{tikz}
\usepackage{comment}
\usepackage{amsmath,amssymb} 
\usepackage{color}

\usepackage{multirow}
\usepackage{colortbl}
\usepackage{float}
\usepackage{marvosym}
\usepackage{ifsym}

\usepackage[accsupp]{axessibility}  

\makeatletter
\def\thanks#1{\protected@xdef\@thanks{\@thanks
        \protect\footnotetext{#1}}}
\makeatother
\begin{document}
\pagestyle{headings}
\mainmatter
\def\ECCVSubNumber{6078}  

\title{Dynamic Local Aggregation Network with Adaptive Clusterer for Anomaly Detection} 

\titlerunning{Dynamic Local Aggregation Network with Adaptive Clusterer}
%
\author{Zhiwei Yang\inst{1}
\and Peng Wu\inst{2}$^\dagger$\thanks{$^\dagger$Corresponding authors.}
\and Jing Liu\inst{1}$^\dagger$
\and 
{Xiaotao Liu\inst{1}}}

\authorrunning{Zhiwei Yang et al.}
%
\institute{Guangzhou Institute of Technology, Xidian University, Guangzhou, China
\and
School of Computer Science, Northwestern Polytechnical University, Xi'an, China 
\email{zwyang97@163.com, xdwupeng@gmail.com, neouma@163.com, xtliu@xidian.edu.cn}}
\maketitle

\begin{abstract} 
Existing methods for anomaly detection based on memory-augmented autoencoder (AE) have the following drawbacks: (1) Establishing a memory bank requires additional memory space. (2) The fixed number of prototypes from subjective assumptions ignores the data feature differences and diversity. To overcome these drawbacks, we introduce \textbf{DLAN-AC}, a \textbf{D}ynamic \textbf{L}ocal \textbf{A}ggregation \textbf{N}etwork with \textbf{A}daptive \textbf{C}lusterer, for anomaly detection. First, The proposed \textbf{DLAN} can automatically learn and aggregate high-level features from the AE to obtain more representative prototypes, while freeing up extra memory space. Second, The proposed \textbf{AC} can adaptively cluster video data to derive initial prototypes with prior information. In addition, we also propose a dynamic redundant clustering strategy (\textbf{DRCS}) to enable DLAN for automatically eliminating feature clusters that do not contribute to the construction of prototypes. Extensive experiments on benchmarks demonstrate that DLAN-AC outperforms most existing methods, validating the effectiveness of our method. Our code is publicly available at https://github.com/Beyond-Zw/DLAN-AC.
\end{abstract}

\section{Introduction}

Unsupervised anomaly detection is to automatically detect events that do not meet our expectations in videos \cite{adam2008robust[9],benezeth2009abnormal[48],chandola2009anomaly[49],sabokrou2015real[50],zhai2016deep[51]}. This is a challenging task, its challenges mainly come from three aspects: (1) There is no clear definition of abnormal behavior, because whether the behavior is abnormal depends on the current environment. (2) Abnormal events rarely occur and cannot be exhaustively listed, which means that it is impossible to collect all abnormal samples. (3) Annotating abnormal frames in a video is an extremely time-consuming task. Due to the limitations of these factors, most current methods only use normal samples for training, and to learn normal behavior patterns. At inference time, samples that are far from the normal pattern are regarded as abnormal. 

The video frame reconstruction and prediction based on autoencoder (AE) thus far are widely used unsupervised video anomaly detection (VAD) paradigms. Such methods have shown promising results, but there are still some limitations. The main limitation is that the generalization ability of AE is so powerful that even some abnormal frames can be reconstructed or predicted well \cite{gong2019memorizing[1]}, \cite{zong2018deep[43]}. Several works are proposed to overcome the above limitation. For example, Gong et al. \cite{gong2019memorizing[1]} and Park et al. \cite{park2020learning[2]} both proposed to record prototypes of normal data by inserting a memory bank into AE, which can enhance the ability of AE to model normal behavior patterns. These two works, to a certain extent, lessen the representation of AE for abnormal video frames, but there are two obvious shortcomings: (1) They need to build a memory bank to store the prototypes of normal data, leading to additional memory overhead. (2) Setting the number of prototypes in the memory bank based on subjective assumptions, which ignores the inherent feature differences in various scenarios, besides, simply setting the same number of prototypes for different datasets limits the features diversity. Recently, Lv et al. \cite{lv2021learning[3]} proposed a meta-prototype network to learn normal dynamics as prototypes. This method learns normal dynamics by constructing a meta-prototype unit that contains an attention mechanism \cite{woo2018cbam[46]}. Using the attention mechanism to learn normal dynamics as prototypes instead of the previous memory bank based method, which can reduce memory overhead. However, high-level semantic information is not excavated due to only using pixel information of feature map. Besides, in this work, setting the number of normal prototypes still based on subjective assumptions.

In this paper, we introduce a novel dynamic local aggregation network with adaptive clusterer (DLAN-AC), which can adaptively cluster video data in different scenarios and dynamically aggregate high-level features to obtain prototypes of normal data. First of all, inspired by the work \cite{arandjelovic2016netvlad[4]}, we propose a dynamic local aggregation network (DLAN), which can automatically learn and aggregate high-level features to obtain more representative prototypes of normal data, and also solve the problem of additional memory consumption. 
Compared with previous memory bank based methods \cite{gong2019memorizing[1],park2020learning[2]}, DLAN can automatically learn prototypes online without additional memory consumptions. Compared with the attention mechanism based method \cite{lv2021learning[3]}, DLAN uses the weighted residual sum between all feature vectors and the cluster center to generate local aggregation features, which can mine higher-level normal features semantic information.

Setting the number of normal prototypes based on subjective assumptions limits the diversity and expressiveness of prototypes. How can we adaptively obtain a reasonable number of prototypes for normal video data according to different scenarios? Inspired by the work \cite{kohonen1990self[6]}, we propose a adaptive clusterer (AC) to adaptively cluster the high-level features, which can provide a prior information for the prototypes setting. Based on AC, we can obtain different cluster numbers and corresponding cluster center vectors for different scenarios, and then use them as the initialization parameters of DLAN. AC not only solves the limitation of manually setting the number of prototypes, but also enables DLAN to be initialized based on the prior information of the data, which can speed up the training process.

In addition, we found that the background features of the high-level feature map are also clustered together by AC, and these background features cannot be used as a representative of the prototypes, because it does not contribute to judging whether the video frame is normal or abnormal. To weaken the influence of these features, drawing inspiration from the work \cite{zhong2018ghostvlad[7]}, we propose a dynamic redundant clustering strategy (DRCS), that is, the final number of clusters is less than the initially set number of clusters in DLAN. Compared with the way of setting a fixed number of redundant clusters in work \cite{zhong2018ghostvlad[7]}, DRCS dynamically adjusts the number of redundant clusters according to the initial number of clusters obtained by AC, which takes into account the difference of the scene. In this way,  DLAN only retain the feature clusters that has important contributions to the establishment of a prototype of normal data.

The main contributions are summarized as follows:
\begin{itemize}
\item We propose a novel dynamic local aggregation network with adaptive clusterer (DLAN-AC), where
DLAN automatically learns and aggregates high-level features to obtain more representative prototypes of the normal video data; AC adaptively clusters the high-level features of video frames according to the scene, which can provide the prior information for prototype setting.
\item We propose a dynamic redundant clustering strategy (DRCS) to eliminate the unimportant feature clusters and retain the feature clusters that has important contributions to the establishment of a prototype of normal video.
\item DLAN-AC is on par with or outperforms other existing methods on three benchmarks, and extensive ablation experiments demonstrate the effectiveness of DLAN-AC.
\end{itemize}

\section{Related Work}

Due to the inherently challenge of the VAD problem, unsupervised learning is the most commonly used method for VAD, where only normal samples are available in the training phase under the unsupervised. To determine whether an abnormal event occurs, a common method is to exploit normal patterns according to their appearance and motion in the training set. Any pattern that is inconsistent with the normal patterns is classified as an anomaly. In the early work \cite{kim2009observe[8],adam2008robust[9],zhao2011online[10],cong2011sparse[11],dutta2015online[12],lu2013abnormal[13],shao2022exploiting[59]}, statistical models and sparse coding are commonly modeling methods that are used in VAD. For example, Adam et al. \cite{adam2008robust[9]} characterized the normal local histograms of optical flows based on statistical monitoring of low-level observations at multiple spatial locations. Zhao et al. in \cite{zhao2011online[10]} proposed a fully unsupervised dynamic sparse coding approach for detecting unusual events in videos based on the online sparse reconstruction. Cong et al. in \cite{cong2011sparse[11]} introduced a sparse reconstruction cost (SRC) over the normal dictionary to measure the normality of testing samples. These early methods have achieved good results in specific scenarios, but due to their poor feature expression ability, the performance is greatly reduced in some complex scenarios. 

In recent years, deep learning has achieved great success in various fields \cite{redmon2016you[15],ren2015faster[16],krizhevsky2012imagenet[17],he2016deep[18],nie2019joint[19],goodfellow2014generative[52],karras2019style[53],shao2022exploiting[59]}. Video anomaly detection methods based on deep learning have begun to emerge widely \cite{hasan2016learning[21],xia2016bottom[22],cai2021appearance[23],georgescu2021anomaly[24],xu2015learning[25],wu2020not[30],wu2019deep[31],liu2021hybrid[33],yang2021bidirectional[39],wang2020cluster[45],zhao2017spatio[42],astrid2021learning[55],astrid2021synthetic[56],9369126[57]}.  For example, Hasan et al. in \cite{hasan2016learning[21]} used deep autoencoder to learn the temporal regularity in the videos for VAD. Luo et al. in \cite{luo2017remembering[27]} presented a convolution autoencoder combined with long short-term memory (LSTM). Recently, Liu et al. in \cite{liu2018future[28]} proposed a future frame prediction framework (FFP) for VAD. A larger difference between the predicted and the actual future frame indicates a possible abnormal event in the frame, which has achieved superior performance over previous reconstruction-based methods. The method of frame reconstruction and frame prediction based on deep learning have shown promising results. However, due to the strong generalization ability of neural networks, some abnormal frames can be also reconstructed or predicted well. This dilemma has caused their performance to encounter bottlenecks. To this end, Gong et al. in ~\cite{gong2019memorizing[1]} embedded a memory module into AE to record the prototypes of normal data, so as to enhance the ability of AE to model normal behavior patterns and weaken the reconstruction of abnormal frames. Park et al. \cite{park2020learning[2]} followed this trend and presented an anomaly detection method that uses multiple prototypes to consider various patterns of normal data, which can obtain more compact and sparse memory bank. Both Gong et al. \cite{gong2019memorizing[1]} and Park et al. \cite{park2020learning[2]} have enhanced the ability of AE to model normal frames by establishing a memory bank. But the disadvantage is that, on the one hand, the memory bank for storing the prototypes of normal data leads to additional memory consumption; on the other hand, the number of prototypes in the memory bank comes from subjective assumptions and is fixed for different datasets, ignoring the scene difference, and limiting the diversity and expressiveness of prototypes. Recently, Lv et al. \cite{lv2021learning[3]} proposed a meta-prototype network to learn normal dynamics as prototypes. This method learns normal dynamics by constructing a meta-prototype unit containing an attention mechanism to reduce the memory overhead. However, there is a lack of mining high-level semantic information of normal features, and the number of prototypes is also based on assumptions. Our method is also targeting at alleviating the excessive generalization ability of AE. However, the difference from the above-mentioned methods is that, on the one hand, our DLAN can automatically learn and aggregate data features to obtain more representative prototypes without additional memory consumption; on the other hand, our AC can adaptively cluster the high-level features of video frames according to the scene, which can provide the prior information for the prototype setting.

\begin{figure}
\centering
\includegraphics[height=5.3cm]{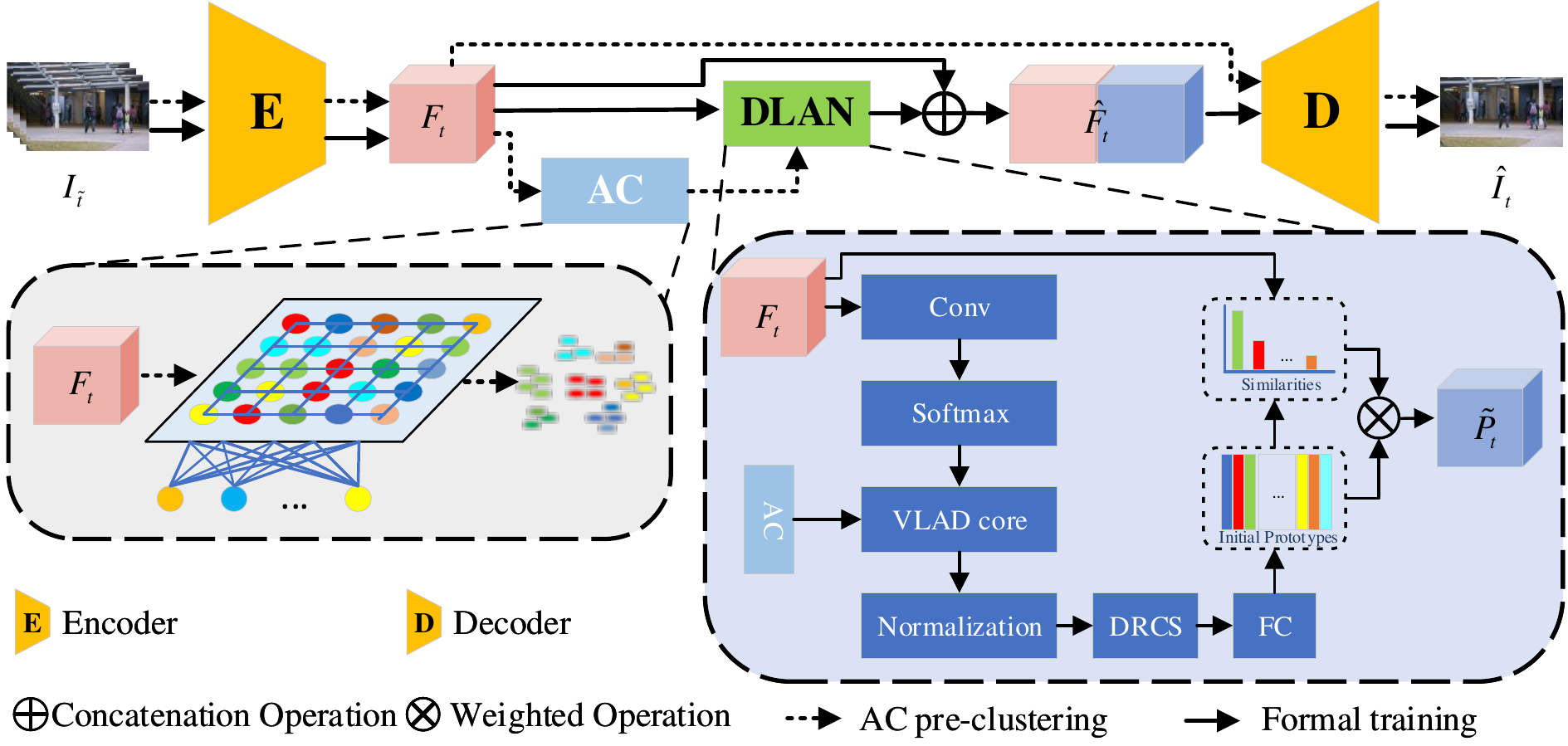}
\caption{An overview of our proposed DLAN-AC. Our method mainly consists of four parts: an encoder, an AC, a DLAN, and a decoder. The encoder extracts a high-level feature map $F_t$ of size $H\times$ $W\times$ $D$ from an input video frame sequence $I_{\tilde t}$. Then, AC adaptively clusters the high-level features to obtain the prior cluster number $M$ and cluster center vector $c_l^*$, which are used to initialize DLAN. Next, the $F_t$ are further locally aggregated to obtain the prototypes $\tilde{P}_t$ by DLAN. Finally, the $\tilde{P}_t$ and the $F_t$ are fused and fed to the decoder to predict the fifth frame $\hat{I}_t$. (Best viewed in color.)}
\label{fig:main}
\end{figure}

\section{Method}
Our method can be divided into four parts: an encoder, an AC, a DLAN, and a decoder. We follow the popular prediction paradigm in the filed of VAD as the main line of the method. First, the encoder extracts high-level features of four consecutive video frames from the input. Then, AC adaptively clusters the high-level features to obtain the prior cluster number and cluster center vector, which are used to initialize DLAN. Then, DLAN further locally aggregates the high-level features to obtain the prototypes of normal video frames. Finally, the prototypes of normal video frames and the original high-level features are fused and fed to the decoder to predict the fifth frame. The overall architecture of DLAN-AC is shown in Fig.~\ref{fig:main}.

\subsection{Encoder and Decoder}
We use AE, a classical framework for reconstruction and prediction tasks, as the main framework of our method. AE is composed of an encoder and a decoder. Encoder is used to extract the high-level features of input video frames, which are processed by AC and DLAN, and then fed to the decoder to predict future frames. Since we follow the prediction paradigm, in order to preserve the background information of the video frame, we use a shortcut connection similar to the U-Net network. Here, we denote by $I_{\tilde t}$ and $F_t$ a video frame sequence and a corresponding high-level feature map from the encoder at time $t$, respectively. The encoder gets the input $I_{\tilde t}$ and outputs $F_t$ of size $H\times$ $W\times$ $D$, where $H$, $W$, $D$ are the height, width, and number of channels, respectively. $F_t^n\in\mathbb{R}^D$ ($n=1,2...N$), where $N=H\times$ $W$, denotes individual high-level feature vector of size $1\times$ $1\times$ $D$ in $F_t$.
\subsection{Adaptive Clusterer}
The main function of AC is to pre-cluster the high-level features extracted by the encoder. The core of AC is Self-Organizing Maps (SOM) neural network. SOM neural network is an unsupervised self-learning neural network, which has the characteristics of good self-organization and easy visualization. It can perform adaptive clustering by identifying the characteristics and inherent relationships between the samples. AC is a two-layer structure consisting of an input layer and output layer (competitive layer). Here, The input layer has $N=H\times$ $W$  neurons in total, and output layer has $L$ neurons. The two layers are connected together by means of a full connection. When AC receives an input vector, the neurons in the output layer compete for the response opportunity to the input vector, and finally the neuron closest to the input vector wins. The weight of the wining neuron and other neurons in its neighborhood will be updated. After multiple rounds of updating, each neuron in the competition layer is most sensitive to a certain type of input vector, so that the associated data in the input layer can be clustered through the competition of neurons in the output layer.

Next, we describe in detail the process of AC clustering high-level features. First, we denote by $c_l\in\mathbb{R}^D$ ($l=1,...L$) the initial weight vector of neurons in the output layer. Then, for each input vector $F_t^n$ , we calculate the Euclidean distance between it and each neuron in the output layer, and find the wining neuron that is the closest to the input vector. The distance calculation formula is as follows:
\begin{equation}
  d_l = \sqrt{\sum_{j=1}^D (F_t^n(j)-c_l(j))^2},
  \label{eq:eq1}
\end{equation}
here, $F_t^n$ is the $n$-th feature in the $F_t$, and $c_l$ is the weight vector of the $l$-th neuron in the output layer. Then, we determine a neighborhood range of the winning neuron, which is generally determined by the neighborhood function. Here the neighborhood function $h(d_n,d_l)$ uses the Gaussian function, and is calculated as follows:
\begin{equation}
h(d_n,d_l) =
\begin{cases}
exp(-\frac{{\left\|d_n-d_l\right\|}^2}{2\delta(t)^2})  & d_n-d_l\leq\delta(t) \\
0 & d_n-d_l>\delta(t)
\end{cases},
\label{eq:eq2}
\end{equation}
here, $\delta(t)=\delta/(1+t/(k/2))$ represents the neighborhood radius, which gradually decreases as time $t$ increasing. $\delta$ is the initial neighborhood radius and $k$ is the number of iterations for each feature map $F_t$. Next, we adjust the weight vector of winning neuron and its neighborhood to move it closer to the input vector. The weight vector is updated as follows:
\begin{equation}
  c_l(t+1) = c_l(t)+\eta(t)h(d_n,d_l)((F_t^n)-c_l(t)),
  \label{eq:eq3}
\end{equation}
where $\eta(t)=\eta/(1+t/(k/2))$ represents the learning rate and $\eta$ is the initial learning rate. This update process is repeated until the network converges. Finally, we obtain the updated neuron weight vector $c_l^*\in\mathbb{R}^D$ ($l=1,...L$) in the output layer, and the value $M$ of the number of clusters obtained after AC clustering. The value $M$ is expressed follows:
\begin{equation}
M=\frac{\begin{matrix} \sum_{i=1}^Q z(F_t) \end{matrix}}{Q},
  \label{eq:eq4}
\end{equation}
here, $z(F_t^n)$ represents the number of wining neurons after responding for each high-level feature map $F_t$ in the AC competitive layer, and $Q$ is the total number of samples in the input dataset.
\subsection{Dynamic Local Aggregation Network}
The Dynamic Local Aggregation Network (DLAN) is mainly composed of convolutional neural networks and a vlad \cite{jegou2010aggregating[5]} core, which can automatically learn and aggregate high-level features to obtain more representative prototypes of normal data, and also solve the problem of additional memory consumption. Specifically, we use the number of clusters $M$ and the cluster center vector $c_l^*$ output by AC as the initialization parameters of DLAN. Then DLAN automatically optimizes the location of these cluster centers and the weight of each feature vector in the $F_t$ to the class center to which it belongs. Next, we calculate the weighted residual sum between each feature vector and the cluster center as the final video frame description vector. These description vectors are then transformed to represent prototypes. In addition, in order to enable DLAN learn to eliminate the insignificant feature clusters and retain the feature clusters that has important contributions to the establishment of a prototype of normal video data, we set dynamic redundant clustering items in DLAN, that is, the final cluster number is less than the initial number of clusters. Here, the initial number of clusters is $L$, namely, the number of neurons in the output layer of AC, and the number of clusters $M$ obtained after AC clustering training is used as the final number of clusters in DLAN. The redundant cluster number is the difference $G=L-M$.

Specifically, the $F_t$ are input into DLAN, and the weighted residual sum of the $N$ feature points in $F_t$ and the cluster center $c_l^*$ is calculated as the element of the local feature matrix $V$. Next, we select the first $M$ elements in the local feature matrix $V$, convert them into a vector form and normalize them to obtain the initial prototypes $P_m\in\mathbb{R}^D$ ($m=1,...M$) by a fully connected layer. The calculation formula of matrix $V$ is as follows:
\begin{equation}
  V(j,l)=\sum_{n=1}^N \beta_l(F_t^n)(F_t^n(j)-c_l^*(j)),~0<{j}\leq D,~0<{l}\leq L,
  \label{eq:eq5}
\end{equation}
where $\beta_l(F_t^n)$ is the contribution weight of each feature point  $F_t^n$ to the $l$-th cluster center $c_l^*$. The calculation formula of $\beta_l(F_t^n)$  is as follows:
\begin{equation}
\beta_l(F_t^n)=\frac{e^{-\alpha{\left\|{F_t^n-c_l^*}\right\|^2}}}{{\begin{matrix}\sum_{l^{'}}{e^{-\alpha\left\|{F_t^n-c_{l^{'}}^*}\right\|^2}}\end{matrix}}}=\frac{e^{{W_l^T F_t^n}+b_l}}{\begin{matrix}\sum_{l^{'}}e^{W_l^T{F_t^n+b_{l^{'}}}}\end{matrix}}.
  \label{eq:eq6}
\end{equation}
Here, following \cite{arandjelovic2016netvlad[4]}, we replace the execution of vlad \cite{jegou2010aggregating[5]} with the convolutional neural network. Where $W_l=2\alpha c_l^*$, $b_l=-\alpha{\left\|c_l^*\right\|^2}$, $\alpha$ is a positive constant that controls the decay of the response with the magnitude of the distance. 
Next, we fuse $F_t$ and $P_m$ to enable the decoder to predict the future frame using normal prototypes, which enhances the ability of AE to model normality and is the key to mitigating the overgeneralization ability of AE for anomalies. First, we calculate the cosine similarity of each $F_t^n$ and the initial prototypes $P_m$, and obtain a 2-dimensional similarity matrix of size $M\times$$N$. Then apply a softmax function to obtain the matching probability weight vector $w_t^{n,m}$ as shown below:
\begin{equation}
w_t^{n,m}=\frac{exp((P_m)^T)F_t^n}{\begin{matrix} \sum_{m^{'}=1}^M exp((P_{m^{'}})^T F_t^n) \end{matrix}}.
  \label{eq:eq7}
\end{equation}
For the initial prototypes $P_m$ , we multiply it with the corresponding matching probability weight $w_t^{n,m}$  to obtain the feature $\tilde{P_t^n}\in\mathbb{R}^D$ as follows: 
\begin{equation}
\tilde{P}_t^n=\sum_{m^{'}=1}^M w_t^{n,{m^{'}}}P_{m^{'}}.
  \label{eq:eq8}
\end{equation}
For each $F_t^n$, we perform Eq.~(\ref{eq:eq7}) and Eq.~(\ref{eq:eq8}) to obtain the final prototypes $\tilde{P}_t\in\mathbb{R}^{H\times W \times D}$. Finally, $\tilde{P}_t$ and $F_t$ are concatenated on the channel to obtain the final fusion feature  ${\hat{F}_t}=F_t\cup\tilde{P}_t$, which is then input to decoder to predict a future frame $\hat{I}_t$.

\subsection{Loss Functions}
 We use intensity loss and gradient loss to force AE to extract the correct high-level features of the input frame sequence, which enable the prototypes learning for normal features representation. In addition, in order to make prototypes have the characteristics of compactness and diversity, we follow the work \cite{park2020learning[2]} using feature compaction loss and feature separation loss to constrain prototypes. 

\noindent\textbf{Intensity loss.} We use the L2 distance to constrain the intensity difference between the predicted frame  $\hat{I}_t$ and real frame $I_t$ :
\begin{equation}
L_{int}=\left\|{\hat{I}_t}-I_t\right\|_2.
  \label{eq:eq9}
\end{equation}

\noindent\textbf{Gradient loss.} To improve the sharpness of the predicted image, we use gradient loss to penalize the gradient difference between the predicted frame and the real frame, where the gradient is the intensity difference between adjacent pixels in the image. The gradient loss function is given as follows:
\begin{equation}
\begin{split}
L_{gd}=\sum_{i,j} \left\|{\left|{I_{i,j}-I_{i-1,j}}\right|} - {\left|{\hat{I}_{i,j}-\hat{I}_{i-1,j}}\right|}\right\|_1 + {\left\|{\left|I_{i,j}-I_{i,j-1}\right| - \left|{{\hat{I}_{i,j}}-{\hat{I}_{i,j-1}}}\right|}\right\|_1},
\end{split}
  \label{eq:eq9}
\end{equation}
where $i,j$ denote the spatial index of a video frame.

\noindent\textbf{Compaction loss.} The feature compactness loss urges the $F_t$ extracted by encoder to approach the prototypes, making the prototypes more compact and reducing intra-class variations, which penalizes the discrepancies between the high-level features and their closest prototype in terms of the L2 norm as: 
\begin{equation}
L_{cp}=\sum_t^T \sum_n^N \left\|{F_t^n-P_a}\right\|_2,
  \label{eq:eq11}
\end{equation}
here $a$ is an index of the item with the greatest probability of matching between $F_t$ and $P_m$:
\begin{equation}
a=\mathop{\arg\max}\limits_{m\in M} w_t^{n,m}.
\label{eq:eq12}
\end{equation}
\noindent\textbf{Separation loss.} The prototypes of normal video frames should be diverse, so these prototypes should be far away from each other. To this end, we use a triplet loss function, which is defined as follows:
\begin{equation}
L_{sp}=\sum_t^T \sum_n^N \left[\left\|{F_t^n-P_a}\right\|_2-\left\|{F_t^n-P_b}\right\|_2+\gamma\right]_+,
\label{eq:eq13}
\end{equation}
here, high-level feature $F_t^n$, the item with the largest matching probability $P_a$, and the item with the second largest matching probability $P_b$, denote the anchor frame, positive example, and hard negative sample, respectively and the margin is denoted by $\gamma$. Similar to  Eq.~(\ref{eq:eq12}), $b$ is expressed as follows:
\begin{equation}
b=\mathop{\arg\max}\limits_{m\in M, m\ne a} w_t^{n,m}.
\label{eq:eq14}
\end{equation}
\noindent\textbf{Overall loss.} The above four loss terms are balanced with $\lambda_{int}$, $\lambda_{gd}$, $\lambda_{cp}$, and $\lambda_{sp}$ as the overall loss function:
 \begin{equation}
L_{all}=\lambda_{int}L_{int}+\lambda_{gd}L_{gd}+\lambda_{cp}L_{cp}+\lambda_{sp}L_{sp}.
  \label{eq:eq15}
\end{equation}
\subsection{Anomaly Detection in Testing Data}
After training on a large number of normal samples, DLAN-AC learns to automatically capture the normal prototype features, so it can predict well for normal frames. For abnormal frames, DLAN-AC cannot capture the abnormal features, and only excavates the normal feature parts, so there will be larger prediction errors. Therefore, we can perform anomaly detection based on the prediction error. Following the work in \cite{mathieu2015deep[29]}, we compute the $PSNR$ between predicted frame $\hat{I}$ and its ground truth $I$ to evaluate the quality of the predicted frame:
\begin{equation}
PSNR(I,\hat{I})=10log_{10}\frac{[{max_{\hat{I}}]^2}}{\frac{1}{K} \begin{matrix}\sum_{i=0}^K(I_i-\hat{I}_i)^2 \end{matrix}},
\label{eq:eq16}
\end{equation}
where $K$ is the total number of image pixels and $max_{\hat{I}}$ is the maximum value of image pixels. The smaller value of $PSNR$ is, the higher probability that the test frame has abnormal behavior, and vice versa. In order to further quantify the probability of anomalies occurring, we also normalize each $PSNR$, following the work in \cite{mathieu2015deep[29]}, to obtain an anomaly score  $S(t)$ in the ranges of [0, 1] as follows:
\begin{equation}
S(t)=1-\frac{PSNR(I_t,\hat{I}_t)-min_t PSNR(I_t,\hat{I}_t)}{max_t PSNR(I_t,\hat{I}_t)-min_t PSNR(I_t,\hat{I}_t)}.
\label{eq:eq17}
\end{equation}

\section{Experiments}
\subsection{Experimental setup}
\noindent\textbf{Datasets.} We evaluate the performance of our method on three benchmarks, which are the most commonly used in the field of VAD. (1) The UCSD Ped2 datasets \cite{li2013anomaly[54]} contains 16 training videos and 12 testing videos with 12 abnormal events, including riding a bike and driving a vehicle on the sidewalk. (2) The CUHK Avenue datasets \cite{lu2013abnormal[13]} consists of 16 training videos and 21 testing videos with 47 abnormal events such as loitering, throwing stuff, and running on the sidewalk. (3) The ShanghaiTech dataset \cite{luo2017revisit[32]} contains 330 training videos and 107 testing videos with 130 abnormal events, such as affray, robbery, and fighting, etc., distributed in 13 different scenes.

\noindent\textbf{Evaluation metric.} Following prior works  \cite{gong2019memorizing[1],liu2018future[28],luo2017revisit[32]}, we evaluate the performance of our proposed method using the Receiver Operation Characteristic curve and the Area Under the Curve (AUC). We use the frame-level AUC metrics for performance evaluation to ensure comparability between different methods. 

\noindent\textbf{Training details.} The training process of our proposed method contains AC pre-clustering and formal training, and the whole process can be executed end-to-end. Firstly, the size of each video frame is resized to 256 $\times$ 256 and the all pixels value are normalized to [-1, 1]. The height $H$ and width $W$ of the high-level feature map, and the numbers of feature channels $D$ are set to 32, 32, and 512, respectively.  During the AC pre-clustering, only the AE and AC module participate. 
For the AE, we use the adam optimizer with an initial learning rate of 2e-4 and decay them using a cosine annealing method. The loss function only contains intensity loss and gradient loss. For AC, the $L$ is set to 25, the $\delta$ and the $\eta$ are set to 0.5, and the $k$ is set to 5000. In this stage, training epochs are set to 10, 10, 5 on Ped2, Avenue and ShanghaiTech, respectively, and the batch size is set to 8. In the formal training stage, AC is shielded and DLAN works. The parameter setting of AE is consistent with the above, but the separation loss and compaction loss are added and the training epochs are set to 80, 80, 20 on Ped2, Avenue and ShanghaiTech, respectively. The weights of the four loss functions are set to $\lambda_{int}=1$, $\lambda_{gd}=1$, $\lambda_{cp}=0.01$, and $\lambda_{sp}=0.01$, and the margin $\gamma=1$. For experimental environment and hyperparameter selection, please see the supplementary materials.
\setlength{\tabcolsep}{4pt}
\begin{table}[t]
\begin{center}
\caption{Quantitative comparison with the state of the art for anomaly detection. We measure the average AUC (\%) on Ped2 \cite{li2013anomaly[54]}, Avenue \cite{lu2013abnormal[13]}, and ShanghaiTech \cite{luo2017revisit[32]}. The comparison methods are listed in chronological order. ('R.' and 'P.' indicate the reconstruction and prediction tasks, respectively.)}
\label{table:tab1}
\begin{tabular}{c|l|c|c|c} 
\hline
\multicolumn{2}{c|}{~~~~~~~~Methods}       & Ped2   & Avenue & ShanghaiTech  \\ 
\hline
\multirow{6}{*}{\rotatebox{90}{—}}       & MPPCA  \cite{kim2009observe[8]}         & 69.3\% & N/A    & N/A       \\
                         & MPPC+SFA \cite{lu2013abnormal[14]}      & 61.3\% & N/A    & N/A       \\
                         & Unmasking  \cite{tudor2017unmasking[34]}     & 82.2\% & 80.6\% & N/A       \\
                         & AMC  \cite{nguyen2019anomaly[35]}           & \underline{96.2\%} & \textbf{86.9\%} & N/A       \\
                         & AnomalyNet \cite{zhou2019anomalynet[36]}  & 94.9\% & 86.1\% & N/A       \\
                         & DeepOC  \cite{wu2019deep[31]}         & \textbf{96.9\%} & \underline{86.6\%} & N/A       \\ 
\hline
\multirow{8}{*}{\rotatebox{90}{R.}}      & Conv-AE \cite{hasan2016learning[21]}  & 90.0\% & 70.2\% & 60.9\%    \\
                         & ConvLSTM-AE \cite{luo2017remembering[27]}    & 88.1\% & 77.0\% & N/A       \\
                         & Stacked RNN \cite{luo2017revisit[32]}    & 92.2\% & 81.7\% & 68.0\%    \\
                         & CDDA \cite{zhou2019anomalynet[36]}           & \underline{96.5\%} & \underline{86.0\%} & 73.3\%    \\
                         & MemAE \cite{gong2019memorizing[1]}          & 94.1\% & 83.3\% & 71.2\%    \\
                         & MNAD \cite{park2020learning[2]}           & 90.2\% & 82.8\% & 69.8\%    \\
                         & AMCM \cite{cai2021appearance[23]}           & \textbf{96.6\%} & \textbf{86.6\%} & \underline{73.7\%}    \\
                         & LNRA-P \cite{astrid2021learning[55]}     & 94.77\% & 84.91\% & 72.46\%  \\
                         & LNRA-SF \cite{astrid2021learning[55]}     & \underline{96.50\%} & 84.67\% & \textbf{75.97\%}  \\
\hline
\multirow{7}{*}{\rotatebox{90}{P.}} & FFP \cite{liu2018future[28]}             & 95.4\% & 84.9\% & 72.8\%    \\
                         & AnoPCN \cite{ye2019anopcn[40]}         & 96.8\% & 86.2\% & 73.6\%    \\
                         & IPRAD \cite{tang2020integrating[38]} & 96.3\% & 85.1\% & 73.0\%    \\
                         & MNAD \cite{park2020learning[2]}          & \underline{97.0\%} & 88.5\% & 70.5\%    \\
                         & ROADMAP \cite{wang2021robust[44]}        & 96.3\% & 88.3\% & \textbf{76.6\%}    \\
                         & MPN \cite{lv2021learning[3]}            & 96.9\% & \underline{89.5\%} & 73.8\%    \\
                         & \textbf{DLAN-AC}         & \textbf{97.6\%} & \textbf{89.9\%} & \underline{74.7\%}    \\
\hline
\end{tabular}
\end{center}
\end{table}
\setlength{\tabcolsep}{1.4pt}

\begin{figure}[t]
\centering
\includegraphics[height=3.0cm]{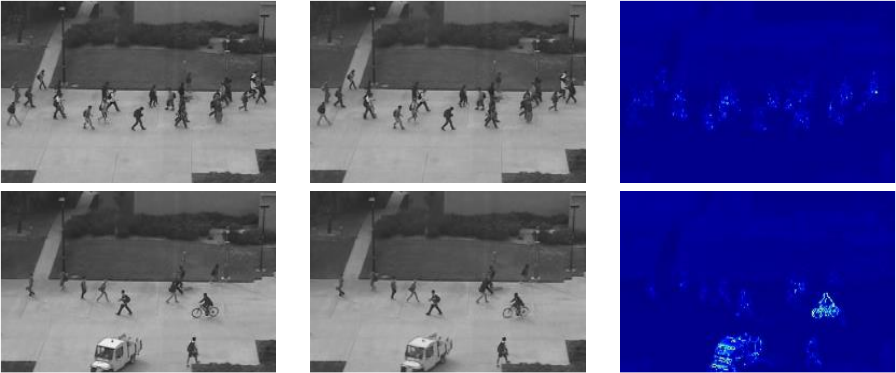}
\caption{Example of frame prediction on Ped2 dataset. The first row is a prediction example of normal event, and the second row is a prediction example of an abnormal event. Left column: the real frame. Mid column: the prediction frame. Right column: the prediction error map. (Best viewed in color.)}
\label{fig:Error_map}
\end{figure}
\begin{figure}[t]
\centering
\includegraphics[height=1.8cm]{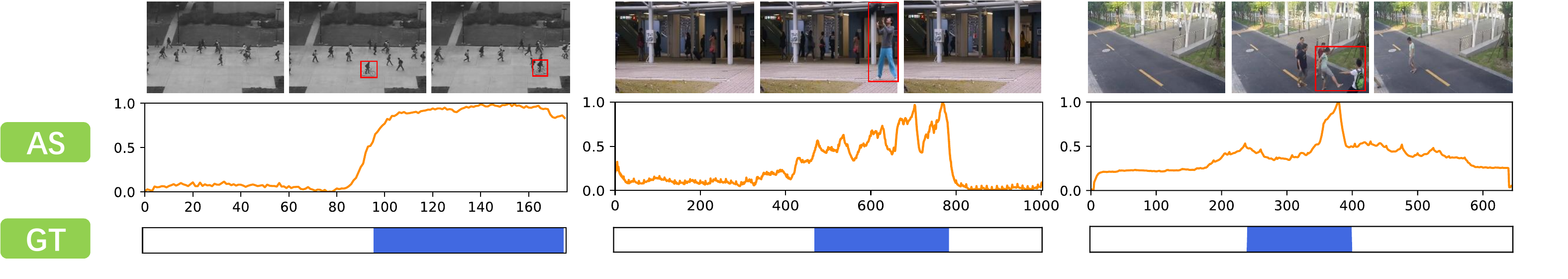}
\caption{Anomaly score curves of several test video clips of our method on three benchmark datasets. AS represents the anomaly score and GT represents the ground truth anomalous frame. (Best viewed in color.)}
\label{fig:AS}
\end{figure}

\subsection{Experimental results}
\noindent\textbf{Comparison with existing methods.} In Table~\ref{table:tab1}, we compare the performance of our method with that of the state-of-the-art methods. It can be seen from Table~\ref{table:tab1} that our method has very strong competitiveness under the comparison of three different levels, almost surpassing most of them. In addition, MemAE \cite{gong2019memorizing[1]}, MNAD \cite{park2020learning[2]}, and MPN \cite{lv2021learning[3]}, are most-related methods to our approach. They build a memory bank for storing prototypes to enhance the ability of AE to model normal behavior patterns and weaken the reconstruction of abnormal frames, but the memory bank requires additional memory storage space. Our method automatically learns to aggregate its normal pattern features directly through the neural network, which frees up extra memory space. Furthermore, the performance of our proposed method in terms of AUC on three benchmarks outperforms that of these three methods, which demonstrates the effectiveness of our method.
\setlength{\tabcolsep}{4pt}
\begin{table}[t]
\begin{center}
\caption{The AUC obtained by DLAN-AC with or without AC on Ped2 \cite{li2013anomaly[54]}, Avenue \cite{lu2013abnormal[13]} and ShanghaiTech \cite{luo2017revisit[32]} dataset. ($M$ stands for the number of prototypes)}
\label{table:tab2}
\begin{tabular}{l!{\color{black}\vrule}c!{\color{black}\vrule}c!{\color{black}\vrule}c} 
\hline
\multirow{2}{*}{~}                                   & Ped2      & Avenue    & ShanghaiTech   \\ 
\cline{2-4}
                                                     & $M$/AUC    & $M$/AUC    & $M$/AUC     \\ 
\hline
\multicolumn{1}{c!{\color{black}\vrule}}{~w/o AC} & 10/97.1\% & 10/89.2\% & 10/74.0\%  \\ 
\hline
\multicolumn{1}{c!{\color{black}\vrule}}{~w AC}   & 13/97.6\% & 13/89.9\% & 11/74.7\%  \\
\hline
\end{tabular}
\end{center}
\end{table}
\setlength{\tabcolsep}{1.4pt}

\noindent\textbf{Qualitative results.} Fig.~\ref{fig:Error_map} shows the future frame prediction results of our method and the corresponding prediction error map for normal events and abnormal events. Obviously, for normal events, the future frame predicted by DLAN-AC is almost close to the actual frame, as shown by the darker error map. For abnormal events, the predicted future frame tends to be blurred and distorted compared to the real frame, and the location of the abnormality is very conspicuous in the error map. In addition, we show the fluctuations of the abnormal scores of several test videos in Fig.~\ref{fig:AS} to illustrate the effectiveness of our method for timely detection of abnormalities. It is easy to observe from Fig.~\ref{fig:AS} that the low abnormal score increases sharply with the occurrence of abnormal events, and then returns to the low level after the abnormality ends. For more qualitative results, please see the supplementary materials.

\subsection{Ablation studies}
In this subsection, we conduct several ablation experiments to analyze the role played by each component of our method.

\begin{figure}[t]
\centering
\includegraphics[height=2.9cm]{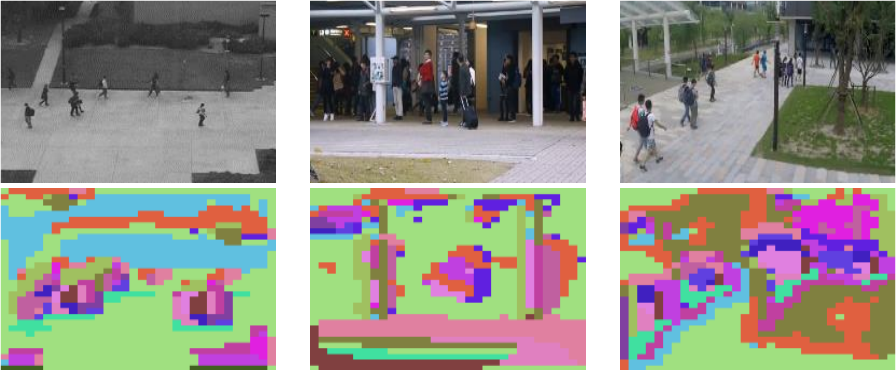}
\caption{AC pre-clustering results on the Ped2, Avenue, and ShanghaiTech. The first row is the original image, and the second row is the corresponding clustering result. It can be seen that elements with similar attributes are grouped into the same category.}
\label{fig:SOFMC_Cluster}
\end{figure}

\noindent\textbf{AC analysis.} To demonstrate the effectiveness of AC, we compare the performance differences between our method using AC to adaptively obtain the number of prototypes and using the pre-set fixed number of prototypes on  three benchmarks in Table~\ref{table:tab2}. The fixed number of setting following \cite{lv2021learning[3]}, which is taken as 10.
It can be seen from Table~\ref{table:tab2} that the performance of the model with the AC is better than that of the model with a fixed number of settings. For different datasets, the number of prototypes $M$ is different, which confirms the inherent difference of video data in different scenes. In addition, we respectively show the cluster distribution map of the high-level features obtained from above three datasets after AC clustering in Fig.~\ref{fig:SOFMC_Cluster}. Taking Ped2 dataset as an example, it can be seen from Fig.~\ref{fig:SOFMC_Cluster} that grass, ground, walls, and people are clearly grouped into different categories. This shows that the adaptive clustering of video features by the AC plays a key role during the pre-clustering of our method, and is an important contributor to the prototype setup of normal data.

\noindent\textbf{DLAN analysis.} In order to analyze the role of DLAN, we freeze the AC, and then set the number of central weight vectors to 10 in DLAN without using DRCS, and initialize them randomly. Next, we insert DLAN into AE as an independent module, and compare it with the AE without DLAN. The comparison results are shown in Table~\ref{table:tab3}. It can be seen that even if there is no other module assistance, just inserting DLAN can increase the performance on the Ped2, Avenue and ShanghaiTech datasets by 2\%, 5.3\%, and 7.3\%, respectively, and DLAN is completely online learning prototypes without consuming additional memory space. This fully proves the effectiveness of using DLAN for feature aggregation to build normal data prototypes. 
\setlength{\tabcolsep}{4pt}
\begin{table}[h]
\begin{center}
\caption{The AUC on Ped2, Avenue and ShanghaiTech datasets with only AE or AE+DLAN.}
\label{table:tab3}
\begin{tabular}{c!{\color{black}\vrule}c!{\color{black}\vrule}c!{\color{black}\vrule}c} 
\hline
\multicolumn{1}{l!{\color{black}\vrule}}{~} & Ped2   & Avenue & ShanghaiTech  \\ 
\hline
AE                                          & 95.1\% & 83.9\% & 66.7\%        \\ 
\hline
AE+DLAN                                     & 97.1\% & 89.2\% & 74.0\%        \\
\hline
\end{tabular}
\end{center}
\end{table}
\setlength{\tabcolsep}{1.4pt}

\begin{figure}[t]
\centering
\includegraphics[height=2.0cm]{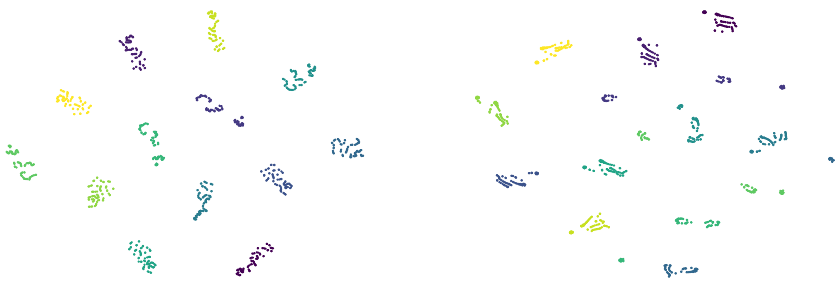}
\caption{t-SNE \cite{van2014accelerating[41]} visualization for the prototypes aggregated by DLAN in the 10th training epoch from Ped2 dataset. The left is the cluster center vector output by AC as the initial clustering vector, and the right is the method of random initialization.}
\label{fig:DLAN_Aggregation}
\end{figure}

\noindent\textbf{Initialization of DLAN.} In order to evaluate the effectiveness of using the cluster center vector output by AC as the initial cluster center vector of DLAN, we respectively compare the aggregation results of using the randomly initialized cluster center vector and the cluster center vector output by AC as the initialization vector. Fig.~\ref{fig:DLAN_Aggregation} visualizes the aggregation results after DLAN aggregation when training to the tenth epoch. It can be clearly seen that DLAN using the cluster center vector output by AC as the initial cluster center has a more compact feature aggregation under the same training epoch. This shows that the initial clustering center vector with certain prior information can make the network convergent faster than the random initialization method.
\setlength{\tabcolsep}{4pt}
\begin{table}[h]
\begin{center}
\caption{The AUC obtained by DLAN-AC with or without Dynamic Redundant Clustering Strategy (DRCS) on Ped2, Avenue and ShanghaiTech datasets.}
\label{table:tab4}
\begin{tabular}{c!{\color{black}\vrule}c!{\color{black}\vrule}c!{\color{black}\vrule}c} 
\hline
\multicolumn{1}{l!{\color{black}\vrule}}{~} & Ped2   & Avenue & ShanghaiTech \\ 
\hline
DLAN-AC
  (w/o DRCS)                         & 97.2\% & 89.4\% & 74.1\% \\ 
\hline
DLAN-AC
  (w DRCS)                           & 97.6\% & 89.9\% & 74.7\% \\
\hline
\end{tabular}
\end{center}
\end{table}
\setlength{\tabcolsep}{1.4pt}

\noindent\textbf{DRCS analysis.} To verify the effectiveness of the DRCS, we respectively compare the AUC on Ped2 and Avenue datasets with and without the number of redundant clusters. It can be seen from Table~\ref{table:tab4} that the performances of the model with the number of redundant clusters set are all better than that of the unset model, which verify the effectiveness of the DRCS to eliminate insignificant feature clusters and retain the feature clusters that have important contributions to the establishment of prototypes.
\section{Conclusions}
We have introduced a novel dynamic local aggregation network with adaptive clusterer (DLAN-AC) for anoamly detection. To this end, we have proposed a dynamic local aggregation network (DLAN), which can automatically learn and aggregate high-level features from AE to obtain more representative normal data prototypes without consuming additional memory space. To adaptively obtain the initial information of prototypes in different scenarios, we have proposed a adaptive clusterer (AC) to perform initial clustering of high-level features to obtain the pre-cluster number and cluster center vector, which are used to initialize the DLAN. In addition, we have also proposed a dynamic redundant clustering strategy (DRCS) to enable DLAN to automatically eliminate the insignificant feature clusters. Extensive experiments on three benchmarks demonstrate that our method outperforms most existing state-of-the-art methods, validating the effectiveness of our method for anomaly detection.

~\\
\textbf{Acknowledgments}
This work was supported in part by the Key Project of Science and Technology Innovation 2030 supported by the Ministry of Science and Technology of China under Grant 2018AAA0101302.

\clearpage
%
%
\bibliographystyle{splncs04}
\bibliography{egbib}

\begin{thebibliography}{10}
\providecommand{\url}[1]{\texttt{#1}}
\providecommand{\urlprefix}{URL }
\providecommand{\doi}[1]{https://doi.org/#1}

\bibitem{adam2008robust[9]}
Adam, A., Rivlin, E., Shimshoni, I., Reinitz, D.: Robust real-time unusual
  event detection using multiple fixed-location monitors. IEEE transactions on
  pattern analysis and machine intelligence  \textbf{30}(3),  555--560 (2008)

\bibitem{arandjelovic2016netvlad[4]}
Arandjelovic, R., Gronat, P., Torii, A., Pajdla, T., Sivic, J.: Netvlad: Cnn
  architecture for weakly supervised place recognition. In: Proceedings of the
  IEEE conference on computer vision and pattern recognition. pp. 5297--5307
  (2016)

\bibitem{astrid2021learning[55]}
Astrid, M., Zaheer, M.Z., Lee, J.Y., Lee, S.I.: Learning not to reconstruct
  anomalies. arXiv preprint arXiv:2110.09742  (2021)

\bibitem{astrid2021synthetic[56]}
Astrid, M., Zaheer, M.Z., Lee, S.I.: Synthetic temporal anomaly guided
  end-to-end video anomaly detection. In: Proceedings of the IEEE/CVF
  International Conference on Computer Vision. pp. 207--214 (2021)

\bibitem{benezeth2009abnormal[48]}
Benezeth, Y., Jodoin, P.M., Saligrama, V., Rosenberger, C.: Abnormal events
  detection based on spatio-temporal co-occurences. In: 2009 IEEE conference on
  computer vision and pattern recognition. pp. 2458--2465. IEEE (2009)

\bibitem{cai2021appearance[23]}
Cai, R., Zhang, H., Liu, W., Gao, S., Hao, Z.: Appearance-motion memory
  consistency network for video anomaly detection. In: Proceedings of the AAAI
  Conference on Artificial Intelligence. pp. 938--946 (2021)

\bibitem{chandola2009anomaly[49]}
Chandola, V., Banerjee, A., Kumar, V.: Anomaly detection: A survey. ACM
  computing surveys (CSUR)  \textbf{41}(3),  1--58 (2009)

\bibitem{cong2011sparse[11]}
Cong, Y., Yuan, J., Liu, J.: Sparse reconstruction cost for abnormal event
  detection. In: CVPR 2011. pp. 3449--3456. IEEE (2011)

\bibitem{dutta2015online[12]}
Dutta, J., Banerjee, B.: Online detection of abnormal events using incremental
  coding length. In: Proceedings of the AAAI Conference on Artificial
  Intelligence (2015)

\bibitem{georgescu2021anomaly[24]}
Georgescu, M.I., Barbalau, A., Ionescu, R.T., Khan, F.S., Popescu, M., Shah,
  M.: Anomaly detection in video via self-supervised and multi-task learning.
  In: Proceedings of the IEEE/CVF Conference on Computer Vision and Pattern
  Recognition. pp. 12742--12752 (2021)

\bibitem{gong2019memorizing[1]}
Gong, D., Liu, L., Le, V., Saha, B., Mansour, M.R., Venkatesh, S., Hengel,
  A.v.d.: Memorizing normality to detect anomaly: Memory-augmented deep
  autoencoder for unsupervised anomaly detection. In: Proceedings of the
  IEEE/CVF International Conference on Computer Vision. pp. 1705--1714 (2019)

\bibitem{goodfellow2014generative[52]}
Goodfellow, I., Pouget-Abadie, J., Mirza, M., Xu, B., Warde-Farley, D., Ozair,
  S., Courville, A., Bengio, Y.: Generative adversarial nets. Advances in
  neural information processing systems  \textbf{27} (2014)

\bibitem{hasan2016learning[21]}
Hasan, M., Choi, J., Neumann, J., Roy-Chowdhury, A.K., Davis, L.S.: Learning
  temporal regularity in video sequences. In: Proceedings of the IEEE
  conference on computer vision and pattern recognition. pp. 733--742 (2016)

\bibitem{he2016deep[18]}
He, K., Zhang, X., Ren, S., Sun, J.: Deep residual learning for image
  recognition. In: Proceedings of the IEEE conference on computer vision and
  pattern recognition. pp. 770--778 (2016)

\bibitem{jegou2010aggregating[5]}
J{\'e}gou, H., Douze, M., Schmid, C., P{\'e}rez, P.: Aggregating local
  descriptors into a compact image representation. In: 2010 IEEE computer
  society conference on computer vision and pattern recognition. pp.
  3304--3311. IEEE (2010)

\bibitem{karras2019style[53]}
Karras, T., Laine, S., Aila, T.: A style-based generator architecture for
  generative adversarial networks. In: Proceedings of the IEEE/CVF conference
  on computer vision and pattern recognition. pp. 4401--4410 (2019)

\bibitem{kim2009observe[8]}
Kim, J., Grauman, K.: Observe locally, infer globally: a space-time mrf for
  detecting abnormal activities with incremental updates. In: 2009 IEEE
  conference on computer vision and pattern recognition. pp. 2921--2928. IEEE
  (2009)

\bibitem{kohonen1990self[6]}
Kohonen, T.: The self-organizing map. Proceedings of the IEEE  \textbf{78}(9),
  1464--1480 (1990)

\bibitem{krizhevsky2012imagenet[17]}
Krizhevsky, A., Sutskever, I., Hinton, G.E.: Imagenet classification with deep
  convolutional neural networks. Advances in neural information processing
  systems  \textbf{25},  1097--1105 (2012)

\bibitem{li2013anomaly[54]}
Li, W., Mahadevan, V., Vasconcelos, N.: Anomaly detection and localization in
  crowded scenes. IEEE transactions on pattern analysis and machine
  intelligence  \textbf{36}(1),  18--32 (2013)

\bibitem{liu2018future[28]}
Liu, W., Luo, W., Lian, D., Gao, S.: Future frame prediction for anomaly
  detection--a new baseline. In: Proceedings of the IEEE conference on computer
  vision and pattern recognition. pp. 6536--6545 (2018)

\bibitem{liu2021hybrid[33]}
Liu, Z., Nie, Y., Long, C., Zhang, Q., Li, G.: A hybrid video anomaly detection
  framework via memory-augmented flow reconstruction and flow-guided frame
  prediction. In: Proceedings of the IEEE/CVF International Conference on
  Computer Vision. pp. 13588--13597 (2021)

\bibitem{lu2013abnormal[13]}
Lu, C., Shi, J., Jia, J.: Abnormal event detection at 150 fps in matlab. In:
  Proceedings of the IEEE international conference on computer vision. pp.
  2720--2727 (2013)

\bibitem{lu2013abnormal[14]}
Lu, C., Shi, J., Jia, J.: Abnormal event detection at 150 fps in matlab. In:
  Proceedings of the IEEE international conference on computer vision. pp.
  2720--2727 (2013)

\bibitem{luo2017remembering[27]}
Luo, W., Liu, W., Gao, S.: Remembering history with convolutional lstm for
  anomaly detection. In: 2017 IEEE International Conference on Multimedia and
  Expo (ICME). pp. 439--444. IEEE (2017)

\bibitem{luo2017revisit[32]}
Luo, W., Liu, W., Gao, S.: A revisit of sparse coding based anomaly detection
  in stacked rnn framework. In: Proceedings of the IEEE International
  Conference on Computer Vision. pp. 341--349 (2017)

\bibitem{lv2021learning[3]}
Lv, H., Chen, C., Cui, Z., Xu, C., Li, Y., Yang, J.: Learning normal dynamics
  in videos with meta prototype network. In: Proceedings of the IEEE/CVF
  Conference on Computer Vision and Pattern Recognition. pp. 15425--15434
  (2021)

\bibitem{mathieu2015deep[29]}
Mathieu, M., Couprie, C., LeCun, Y.: Deep multi-scale video prediction beyond
  mean square error. arXiv preprint arXiv:1511.05440  (2015)

\bibitem{nguyen2019anomaly[35]}
Nguyen, T.N., Meunier, J.: Anomaly detection in video sequence with
  appearance-motion correspondence. In: Proceedings of the IEEE/CVF
  International Conference on Computer Vision. pp. 1273--1283 (2019)

\bibitem{nie2019joint[19]}
Nie, X., Jing, W., Cui, C., Zhang, C.J., Zhu, L., Yin, Y.: Joint multi-view
  hashing for large-scale near-duplicate video retrieval. IEEE Transactions on
  Knowledge and Data Engineering  \textbf{32}(10),  1951--1965 (2019)

\bibitem{park2020learning[2]}
Park, H., Noh, J., Ham, B.: Learning memory-guided normality for anomaly
  detection. In: Proceedings of the IEEE/CVF Conference on Computer Vision and
  Pattern Recognition. pp. 14372--14381 (2020)

\bibitem{redmon2016you[15]}
Redmon, J., Divvala, S., Girshick, R., Farhadi, A.: You only look once:
  Unified, real-time object detection. In: Proceedings of the IEEE conference
  on computer vision and pattern recognition. pp. 779--788 (2016)

\bibitem{ren2015faster[16]}
Ren, S., He, K., Girshick, R., Sun, J.: Faster r-cnn: Towards real-time object
  detection with region proposal networks. Advances in neural information
  processing systems  \textbf{28},  91--99 (2015)

\bibitem{sabokrou2015real[50]}
Sabokrou, M., Fathy, M., Hoseini, M., Klette, R.: Real-time anomaly detection
  and localization in crowded scenes. In: Proceedings of the IEEE conference on
  computer vision and pattern recognition workshops. pp. 56--62 (2015)

\bibitem{shao2022exploiting[59]}
Shao, F., Liu, J., Wu, P., Yang, Z., Wu, Z.: Exploiting foreground and
  background separation for prohibited item detection in overlapping x-ray
  images. Pattern Recognition  \textbf{122},  108261 (2022)

\bibitem{tang2020integrating[38]}
Tang, Y., Zhao, L., Zhang, S., Gong, C., Li, G., Yang, J.: Integrating
  prediction and reconstruction for anomaly detection. Pattern Recognition
  Letters  \textbf{129},  123--130 (2020)

\bibitem{tudor2017unmasking[34]}
Tudor~Ionescu, R., Smeureanu, S., Alexe, B., Popescu, M.: Unmasking the
  abnormal events in video. In: Proceedings of the IEEE international
  conference on computer vision. pp. 2895--2903 (2017)

\bibitem{van2014accelerating[41]}
Van Der~Maaten, L.: Accelerating t-sne using tree-based algorithms. The Journal
  of Machine Learning Research  \textbf{15}(1),  3221--3245 (2014)

\bibitem{wang2021robust[44]}
Wang, X., Che, Z., Jiang, B., Xiao, N., Yang, K., Tang, J., Ye, J., Wang, J.,
  Qi, Q.: Robust unsupervised video anomaly detection by multipath frame
  prediction. IEEE Transactions on Neural Networks and Learning Systems  (2021)

\bibitem{wang2020cluster[45]}
Wang, Z., Zou, Y., Zhang, Z.: Cluster attention contrast for video anomaly
  detection. In: Proceedings of the 28th ACM International Conference on
  Multimedia. pp. 2463--2471 (2020)

\bibitem{woo2018cbam[46]}
Woo, S., Park, J., Lee, J.Y., Kweon, I.S.: Cbam: Convolutional block attention
  module. In: Proceedings of the European conference on computer vision (ECCV).
  pp. 3--19 (2018)

\bibitem{9369126[57]}
Wu, P., Liu, J.: Learning causal temporal relation and feature discrimination
  for anomaly detection. IEEE Transactions on Image Processing  \textbf{30},
  3513--3527 (2021). \doi{10.1109/TIP.2021.3062192}

\bibitem{wu2019deep[31]}
Wu, P., Liu, J., Shen, F.: A deep one-class neural network for anomalous event
  detection in complex scenes. IEEE transactions on neural networks and
  learning systems  \textbf{31}(7),  2609--2622 (2019)

\bibitem{wu2020not[30]}
Wu, P., Liu, J., Shi, Y., Sun, Y., Shao, F., Wu, Z., Yang, Z.: Not only look,
  but also listen: Learning multimodal violence detection under weak
  supervision. In: European Conference on Computer Vision. pp. 322--339.
  Springer (2020)

\bibitem{xia2016bottom[22]}
Xia, C., Qi, F., Shi, G.: Bottom--up visual saliency estimation with deep
  autoencoder-based sparse reconstruction. IEEE transactions on neural networks
  and learning systems  \textbf{27}(6),  1227--1240 (2016)

\bibitem{xu2015learning[25]}
Xu, D., Ricci, E., Yan, Y., Song, J., Sebe, N.: Learning deep representations
  of appearance and motion for anomalous event detection. arXiv preprint
  arXiv:1510.01553  (2015)

\bibitem{yang2021bidirectional[39]}
Yang, Z., Liu, J., Wu, P.: Bidirectional retrospective generation adversarial
  network for anomaly detection in videos. IEEE Access  \textbf{9},
  107842--107857 (2021)

\bibitem{ye2019anopcn[40]}
Ye, M., Peng, X., Gan, W., Wu, W., Qiao, Y.: Anopcn: Video anomaly detection
  via deep predictive coding network. In: Proceedings of the 27th ACM
  International Conference on Multimedia. pp. 1805--1813 (2019)

\bibitem{zhai2016deep[51]}
Zhai, S., Cheng, Y., Lu, W., Zhang, Z.: Deep structured energy based models for
  anomaly detection. In: International conference on machine learning. pp.
  1100--1109. PMLR (2016)

\bibitem{zhao2011online[10]}
Zhao, B., Fei-Fei, L., Xing, E.P.: Online detection of unusual events in videos
  via dynamic sparse coding. In: CVPR 2011. pp. 3313--3320. IEEE (2011)

\bibitem{zhao2017spatio[42]}
Zhao, Y., Deng, B., Shen, C., Liu, Y., Lu, H., Hua, X.S.: Spatio-temporal
  autoencoder for video anomaly detection. In: Proceedings of the 25th ACM
  international conference on Multimedia. pp. 1933--1941 (2017)

\bibitem{zhong2018ghostvlad[7]}
Zhong, Y., Arandjelovi{\'c}, R., Zisserman, A.: Ghostvlad for set-based face
  recognition. In: Asian conference on computer vision. pp. 35--50. Springer
  (2018)

\bibitem{zhou2019anomalynet[36]}
Zhou, J.T., Du, J., Zhu, H., Peng, X., Liu, Y., Goh, R.S.M.: Anomalynet: An
  anomaly detection network for video surveillance. IEEE Transactions on
  Information Forensics and Security  \textbf{14}(10),  2537--2550 (2019)

\bibitem{zong2018deep[43]}
Zong, B., Song, Q., Min, M.R., Cheng, W., Lumezanu, C., Cho, D., Chen, H.: Deep
  autoencoding gaussian mixture model for unsupervised anomaly detection. In:
  International conference on learning representations (2018)

\end{thebibliography}
\end{document}